\documentclass[11pt,letterpaper]{article}
\usepackage{emnlp2016}
\usepackage{times}
\usepackage{helvet}
\usepackage{courier}
\usepackage{graphicx}
\usepackage{amsmath}
\usepackage{url}
\usepackage{multirow}
\usepackage{multicol}
\usepackage{tabularx}
\usepackage{arydshln}
\usepackage{latexsym}
\usepackage{amsfonts}
\usepackage{MnSymbol}
\usepackage{subfigure}

\usepackage{tikz}
\usepackage{tikz-dependency}
\usepackage[english]{babel}

\newcommand{\specialcell}[2][c]{%
	\begin{tabular}[#1]{@{}l@{}}#2\end{tabular}}

\newcolumntype{L}[1]{>{\raggedright\arraybackslash}p{#1}}
\newcolumntype{C}[1]{>{\centering\arraybackslash}p{#1}}
\newcolumntype{R}[1]{>{\raggedleft\arraybackslash}p{#1}}

\emnlpfinalcopy

\def\emnlppaperid{***}

\title{Exploiting Multi-typed Treebanks for Parsing with \\Deep Multi-task Learning}

\author{Jiang Guo$^\spadesuit$, Wanxiang Che$^\spadesuit$, Haifeng Wang$^\heartsuit$, Ting Liu$^\spadesuit$ \\
	$^\spadesuit$Center for Social Computing and Information Retrieval, Harbin Institute of Technology, China \\
	$^\heartsuit$Baidu Inc., China \\
	{\tt \{jguo, car, tliu\}@ir.hit.edu.cn}\\
	{\tt wanghaifeng@baidu.com}
}

\date{}

\begin{document}
	
	\maketitle
	
	\begin{abstract}
		
		Various treebanks have been released for dependency parsing.
		Despite that treebanks may belong to different languages or have different annotation schemes, they contain syntactic knowledge that is potential to benefit each other.
		This paper presents an universal framework for exploiting these multi-typed treebanks to improve parsing with deep multi-task learning.
		We consider two kinds of treebanks as source: the \textit{multilingual universal treebanks} and the \textit{monolingual heterogeneous treebanks}.
		Multiple treebanks are trained jointly and interacted with multi-level parameter sharing.
		Experiments on several benchmark datasets in various languages demonstrate that our approach can make effective use of arbitrary source treebanks to improve target parsing models.
		
	\end{abstract}
	
	\section{Introduction}

	As a long-standing central problem in natural language processing (NLP), dependency parsing has been dominated by data-driven approaches with supervised learning for decades.
	The foundation of data-driven parsing is the availability and scale of annotated training data (i.e., \textit{treebanks}).
	Numerous efforts have been made towards the construction of treebanks which established the benchmark research on dependency parsing, such as the widely-used Penn Treebank~\cite{marcus1993building}.
	However, the heavy cost of treebanking typically limits the existing treebanks in both scale and coverage of languages.
	
	To address the problem, a variety of authors have proposed to exploit existing heterogeneous treebanks with different annotation schemes via grammar conversion~\cite{niu-wang-wu:2009:ACLIJCNLP}, quasi-synchronous grammar features~\cite{li-liu-che:2012:ACL2012} or shared feature representations~\cite{johansson:2013:NAACL-HLT} for the enhancement of parsing models.
	Despite their effectiveness in specific datasets, these methods typically require manually designed rules or features, and in most cases, they are limited to the data resources that can be used.
	Furthermore, for the majority of world languages, such heterogeneous treebanks are not even available.
	In these cases, cross-lingual treebanks may lend a helping hand.
	
	
	In this paper, we aim at developing an universal framework that can exploit multi-typed source treebanks to improve parsing of a target treebank.
	Specifically, we will consider two kinds of source treebanks, that are \textit{multilingual universal treebanks} and \textit{monolingual heterogeneous treebanks}.

\begin{figure}[t]
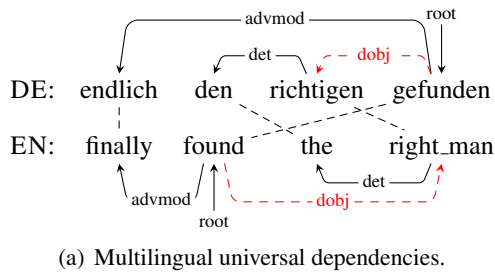
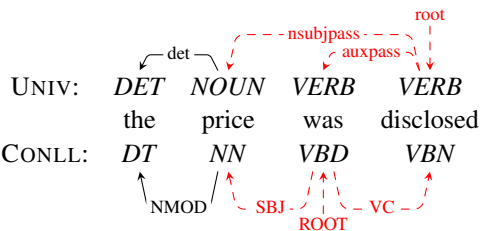

		\centering
		\small
		\subfigure[Multilingual universal dependencies.]{\label{fig:multilingual}
			\begin{dependency}[theme=default] 
				\begin{deptext}[column sep=0.5em]
					\textsc{DE}: \& endlich \& den \& richtigen \& {gefunden} \\[2.5ex]
					\textsc{EN}: \& finally \& {found} \& the \& right\_man \\
				\end{deptext}
				\depedge[edge style={black}, label style={draw=none, text=black}, edge unit distance=1.5ex]{5}{2}{advmod}
				\depedge[edge style={black}, label style={draw=none, text=black}, edge unit distance=1.5ex]{4}{3}{det}
				\depedge[edge style={red, dashed}, label style={draw=none, text=red}, edge unit distance=1.5ex]{5}{4}{dobj}
				\deproot[edge style={black}, label style={draw=none, text=black}, edge unit distance=1.6ex]{5}{root}
				\depedge[edge below, edge style={black}, label style={draw=none, text=black, edge unit distance=1.5ex}]{3}{2}{advmod}
				\depedge[edge below, edge style={black}, label style={draw=none, text=black}, edge unit distance=1.6ex]{5}{4}{det}
				\depedge[edge below, edge style={red, dashed}, label style={draw=none, text=red}, edge unit distance=1.6ex]{3}{5}{dobj}
				\deproot[edge below, edge style={black}, label style={draw=none, text=black}, edge unit distance=1.6ex]{3}{root}
				\draw [black, densely dashed] (\wordref{1}{2})--(\wordref{2}{2});
				\draw [black, densely dashed] (\wordref{1}{3})--(\wordref{2}{4});
				\draw [black, densely dashed] (\wordref{1}{4})--(\wordref{2}{5});
				\draw [black, densely dashed] (\wordref{1}{5})--(\wordref{2}{3});
			\end{dependency}
		}
		\subfigure[Monolingual heterogeneous dependencies.]{\label{fig:heterogeneous}
			\begin{dependency}[theme=default]
				\begin{deptext}[column sep=0.5em, row sep=.1ex]
					\textsc{Univ}: \& \small{\textit{DET}} \& \small{\textit{NOUN}} \& \small{\textit{VERB}} \& \small{\textit{VERB}} \\
					\& the \& price \& was \& disclosed \\
					\textsc{Conll}: \& \small{\textit{DT}} \& \small{\textit{NN}} \& \small{\textit{VBD}} \& \small{\textit{VBN}} \\
				\end{deptext}
				\deproot[edge style={red, dashed}, label style={draw=none, text=red}, edge unit distance=1.5ex]{5}{root}
				\depedge[edge style={black}, label style={draw=none, text=black}, edge unit distance=1.3ex]{3}{2}{det}
				\depedge[edge style={red, dashed}, label style={draw=none, text=red}, edge unit distance=1.5ex]{5}{3}{nsubjpass}
				\depedge[edge style={red, dashed}, label style={draw=none, text=red}, edge unit distance=1.5ex]{5}{4}{auxpass}
				\depedge[edge below, edge style={black}, label style={draw=none, text=black, edge unit distance=1.3ex}]{3}{2}{NMOD}
				\depedge[edge below, edge style={red, dashed}, label style={draw=none, text=red, edge unit distance=1.3ex},]{4}{3}{SBJ}
				\depedge[edge below, edge style={red, dashed}, label style={draw=none, text=red, edge unit distance=1.3ex},]{4}{5}{VC}
				\deproot[edge below, edge style={red, dashed}, label style={draw=none, text=red}, edge unit distance=1.5ex]{4}{ROOT}
			\end{dependency}
		}	
		\caption{Comparisons between multilingual universal dependencies (a) and monolingual heterogeneous dependencies (b).} \label{fig:running-example}
		\vspace{-0.5em}
	\end{figure}

	Cross-lingual supervision has proven highly beneficial for low-resource language parsing~\cite{hwa2005bootstrapping,mcdonald2011multi}, implying that different languages have a great deal of common ground in grammars.
	But unfortunately, linguistic inconsistencies also exist in both typologies and lexical representations across languages.
	Figure~\ref{fig:multilingual} illustrates two sentences in German and English with universal dependency annotations.
	The typological differences (\textit{subject-verb-object} order) results in the opposite directions of the \textit{dobj} arcs, while the rest arcs remain consistent.
	
	Similar problems also come with monolingual heterogeneous treebanks.
	Figure~\ref{fig:heterogeneous} shows an English sentence annotated with respectively the universal dependencies which are \textit{content-head} and the \textsc{CoNLL} dependencies which instead take the functional heads.
	Despite the structural divergences, these treebanks express the syntax of the same language, thereby sharing a large amount of common knowledge that can be effectively transferred.

	The present paper proposes a simple and effective framework that aims at making full use of the consistencies while avoids suffering from the inconsistencies across treebanks.
	Our framework effectively ties together the deep neural parsing models with multi-task learning, using multi-level parameter sharing to control the information flow across tasks.
	More specifically, learning with each treebank is maintained as an individual task, and their interactions are achieved through parameter sharing in different abstraction levels on the deep neural network, thus referred to as \textit{deep multi-task learning}.
	We find that different parameter sharing strategies should be applied for different typed source treebanks adaptively, due to the different types of consistencies and inconsistencies (Figure~\ref{fig:running-example}).
	
	
	
	We investigate the effect of multilingual treebanks as source using the Universal Dependency Treebanks (UDT)~\cite{mcdonald2013universal}.
	We show that our approach improves significantly over strong supervised baseline systems in six languages.
	We further study the effect of monolingual heterogeneous treebanks as source using \textsc{UDT} and the \textsc{CoNLL-X} shared task dataset~\cite{buchholz2006conll}.
	We consider using UDT and CoNLL-X as source treebanks respectively, to investigate their mutual benefits.
	Experiment results show significant improvements under both settings.
	Moreover, indirect comparisons on the Chinese Penn Treebank 5.1 (CTB5) using the Chinese Dependency Treebank (CDT)\footnote{\url{catalog.ldc.upenn.edu/LDC2012T05}} as source treebank show the merits of our approach over previous work.
	
	
	
	
	\section{Related Work}
	\label{sec:related-work}
	
	The present work is related to several strands of previous studies.
	
	\textbf{Monolingual resources for parsing}.
    Exploiting heterogeneous treebanks for parsing has been explored in various ways.
	\newcite{niu-wang-wu:2009:ACLIJCNLP} automatically convert the dependency-structure CDT into the phrase-structure style of CTB5 using a trained constituency parser on CTB5, and then combined the converted treebanks for constituency parsing.
	\newcite{li-liu-che:2012:ACL2012} capture the annotation inconsistencies among different treebanks by designing several types of \textit{transformation patterns}, based on which they introduce \textit{quasi-synchronous grammar} features~\cite{smith-eisner:2009:EMNLP} to augment the baseline parsing models.
	\newcite{johansson:2013:NAACL-HLT} also adopts the idea of parameter sharing to incorporate multiple treebanks.
	They focused on parameter sharing at feature-level with discrete representations, which limits its scalability to multilingual treebanks where feature surfaces might be totally different.
	On the contrary, our approach are capable of utilizing representation-level parameter sharing, making full use of the multi-level abstractive representations generated by deep neural network.
	This is the key that makes our framework scalable to multi-typed treebanks and thus more practically useful.
	
	Aside from resource utilization, attempts have also been made to integrate different parsing models through stacking~\cite{torresmartins-EtAl:2008:EMNLP,nivre-mcdonald:2008:ACLMain} or joint inference~\cite{zhang-clark:2008:EMNLP,zhang-EtAl:2014:Coling2}.
	
	\textbf{Multilingual resources for parsing}.
	Cross-lingual transfer has proven to be a promising way of inducing parsers for low-resource languages, either through \textit{data transfer}~\cite{hwa2005bootstrapping,tiedemann2014rediscovering,rasooli-collins:2015:EMNLP} or \textit{model transfer}~\cite{mcdonald2011multi,tackstrom2012cross,guo-EtAl:2015:ACL-IJCNLP2,zhang2015multi}.
	
	\newcite{duong-EtAl:2015:EMNLP} and~\newcite{waleed2016unidep} both adopt parameter sharing to exploit multilingual treebanks in parsing, but with a few important differences to our work.
	In both of their models, most of the neural network parameters are shared in two (or multiple) parsers except the feature embeddings,\footnote{\newcite{duong-EtAl:2015:EMNLP} used $L2$ regularizers to tie the lexical embeddings with a bilingual dictionary.} which ignores the important \textit{syntactical inconsistencies} of different languages and is also inapplicable for heterogeneous treebanks that have different transition actions.
	Besides, \newcite{duong-EtAl:2015:EMNLP} focus on low resource parsing where the target language has a small treebank of $\sim$ 3K tokens.
	Their models may sacrifice accuracy on target languages with a large treebank.
	\newcite{waleed2016unidep} instead train a single parser on a multilingual set of rich-resource treebanks, which is a more similar setting to ours.
	We refer to their approach as \textit{shallow multi-task learning} (\textsc{SMTL}) and will include as one of our baseline systems (Section~\ref{subsec:baseline}).
	Note that \textsc{SMTL} is a special case of our approach in which all tasks use the same set of parameters.
	
	Bilingual parallel data has also proven beneficial in various ways~\cite{chen-kazama-torisawa:2010:ACL,huang-jiang-liu:2009:EMNLP,burkett-klein:2008:EMNLP}, demonstrating the potential of cross-lingual transfer learning.
	
	\textbf{Multi-task learning for NLP}.
	There has been a line of research on joint modeling pipelined NLP tasks, such as word segmentation, POS tagging and parsing~\cite{hatori-EtAl:2012:ACL2012,li-EtAl:2011:EMNLP3,bohnet-nivre:2012:EMNLP-CoNLL}.
	Most multi-task learning or joint training frameworks can be summarized as parameter sharing approaches proposed by~\newcite{Ando:2005:FLP:1046920.1194905}.
	In the context of neural models for NLP, the most notable work was proposed by~\newcite{collobert2008unified}, which aims at solving multiple NLP tasks within one framework by sharing common word embeddings.
	\newcite{henderson2013multilingual} present a joint dependency parsing and semantic role labeling model with the Incremental Sigmoid Belief Networks (ISBN)~\cite{Henderson:2010:ISB:1756006.1953044}.
	More recently, the idea of neural multi-task learning was applied to sequence-to-sequence problems with recurrent neural networks.
	\newcite{dong-EtAl:2015:ACL-IJCNLP2} use multiple decoders in neural machine translation systems that allows translating one source language to many target languages. 
	\newcite{luong2015multitask} study the ensemble of a wide range of tasks (e.g., syntactic parsing, machine translation, image caption, etc.) with multi-task sequence-to-sequence models.
	
	To the best of our knowledge, we present the first work that successfully integrate both monolingual and multilingual treebanks for parsing, with or without consistent annotation schemes.

	\section{Approach}
	This section describes the deep multi-task learning architecture, using a formalism that extends on the transition-based dependency parsing model with LSTM networks~\cite{dyer-EtAl:2015:ACL-IJCNLP} which is further enhanced by modeling characters~\cite{ballesteros-dyer-smith:2015:EMNLP}.
	We first revisit the parsing approach of~\newcite{ballesteros-dyer-smith:2015:EMNLP}, then present our framework for learning with multi-typed source treebanks.
	
	\subsection{Transition-based Neural Parsing}
	
	Neural models for parsing have gained a lot of interests in recent years, particularly boosted by~\newcite{chen2014fast}.
	The heart of transition-based parsing is the challenge of representing the \textit{state} (configuration) of a transition system, based on which the most likely transition action is determined.
	Typically, a state includes three primary components, a \textit{stack}, a \textit{buffer} and a set of \textit{dependency arcs}.
	Traditional parsing models deal with features extracted from manually defined feature templates in a discrete feature space, which suffers from the problems of \textit{Sparsity}, \textit{Incompleteness} and \textit{Expensive feature computation}.
	The neural network model proposed by~\newcite{chen2014fast} instead represents features as continuous, low-dimensional vectors and use a \textit{cube} activation function for implicit feature composition.
	More recently, this architecture has been improved in several different ways~\cite{dyer-EtAl:2015:ACL-IJCNLP,weiss-EtAl:2015:ACL-IJCNLP,zhou-EtAl:2015:ACL-IJCNLP3,andor2016globally}.
	Here, we employ the LSTM-based architecture enhanced with character bidirectional LSTMs~\cite{ballesteros-dyer-smith:2015:EMNLP} for the following major reasons:
	\begin{itemize}
		\item Compared with Chen \& Manning's architecture, it makes full use of the non-local features by modeling the full history information of a \textit{state} with stack LSTMs.
		\item By modeling words, stack, buffer and action sequence separately which indicate hierarchical abstractions of representations, we can control the information flow across tasks via parameter sharing with more flexibility (Section~\ref{subsec:mtl}).
	\end{itemize}
	
	\begin{figure}[t]
		\centering
		\vspace{-0.5em}
		\includegraphics[width=70mm]{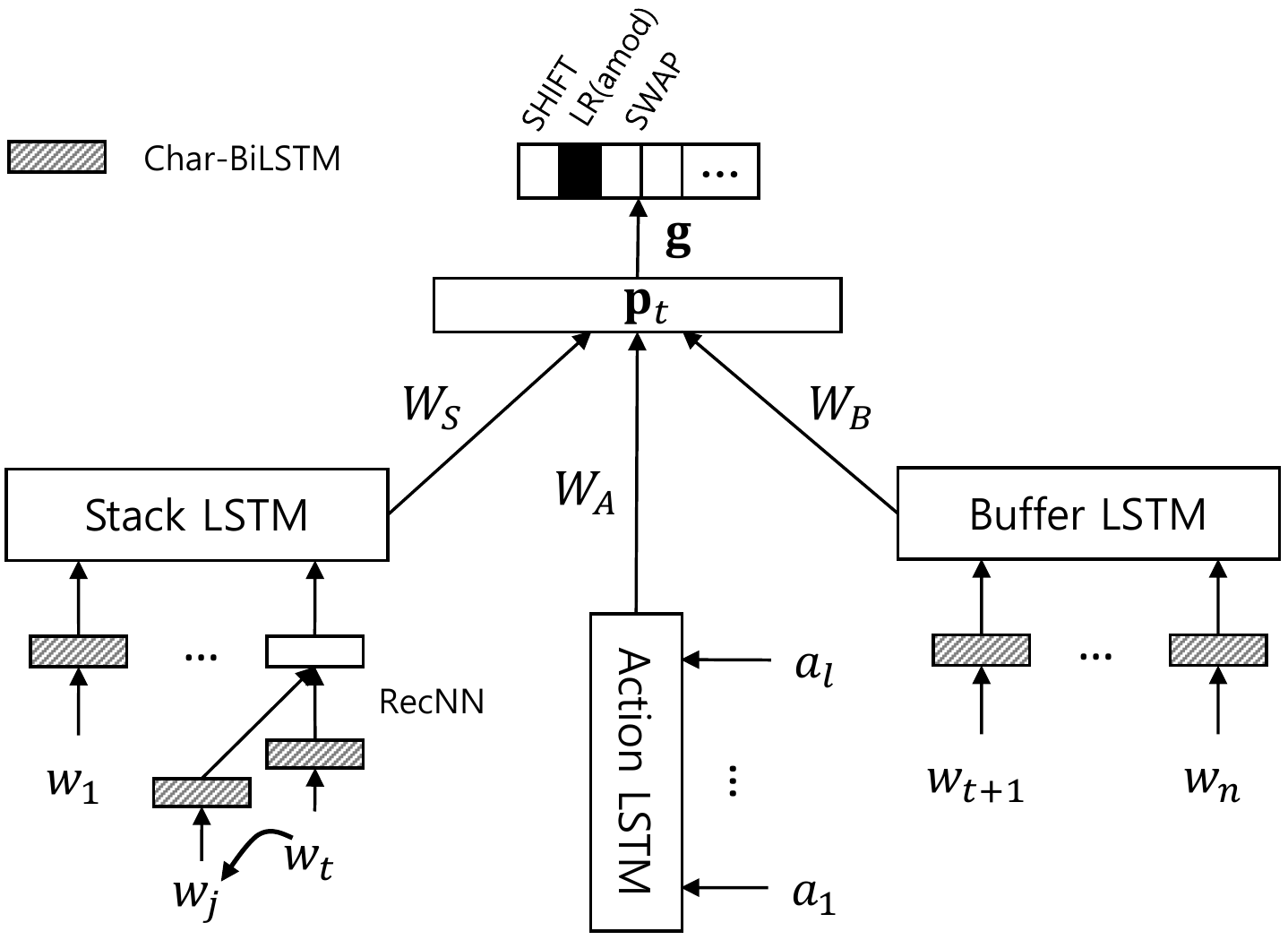}
		\caption{The LSTM-based neural parser.}
		\label{fig:lstm-parser}
	\end{figure}
	\begin{figure}[t]
		\centering
		\includegraphics[width=45mm]{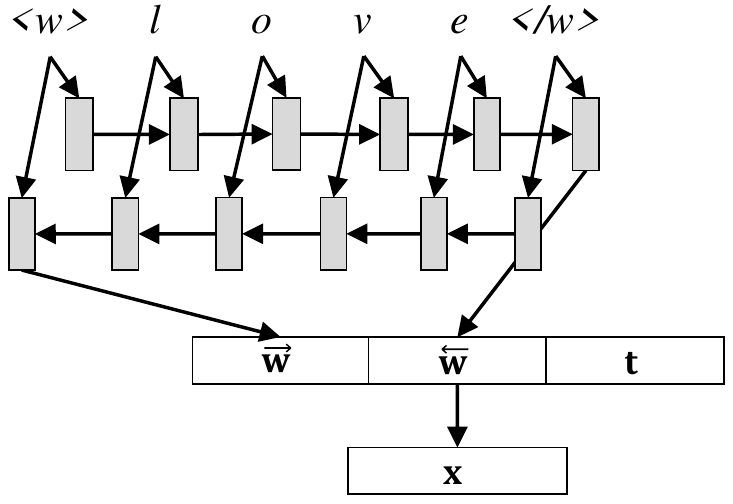}
		\vspace{-0.5em}
		\caption{Char-BiLSTM modeling the embedding of \textit{love}.}
		\label{fig:char-bilstm}
		\vspace{-0.5em}
	\end{figure}
	
	Besides, we did not use the earlier ISBN parsing model~\cite{titov-henderson:2007:EMNLP-CoNLL2007} due to its lack of scalability to large vocabulary.
	Figure~\ref{fig:lstm-parser} illustrates the transition-based parsing architecture using LSTMs.
	Bidirectional LSTMs are used for modeling the word representations (Figure~\ref{fig:char-bilstm}), which we refer to as Char-BiLSTMs henceforth.
	Char-BiLSTMs learn features for each word, and then the representation of each token can be calculated as:
	\begin{equation}
		\mathbf{x} = \mathtt{ReLU}(\mathbf{V}[\overrightarrow{\mathbf{w}}; \overleftarrow{\mathbf{w}}; \mathbf{t}] + \mathbf{b})
	\end{equation}
	where $\mathbf{t}$ is the POS tag embedding.
	The token embeddings are then fed into subsequent LSTM layers to obtain representations of the \textit{stack}, \textit{buffer} and \textit{action sequence} respectively referred to as $\mathbf{s}_t, \mathbf{b}_t$ and $\mathbf{a}_t$ (The subscript $t$ represents the time step).
	Note that the subtrees within the stack and buffer are modeled with a \textit{recursive neural network} (RecNN) as described in~\newcite{dyer-EtAl:2015:ACL-IJCNLP}.
	Next, a linear mapping ($\mathbf{W}$) is applied to the concatenation of $\mathbf{s}_t, \mathbf{b}_t$ and $\mathbf{a}_t$, and passed through a component-wise $\mathtt{ReLU}$:
	\begin{equation}
		\mathbf{p}_t = \mathtt{ReLU}(\mathbf{W}[\mathbf{s}_t; \mathbf{b}_t; \mathbf{a}_t] + \mathbf{d})
		\label{eqn:1}
	\end{equation}
	Finally, the probability of next action $z \in \mathcal{A}(S, B)$ is estimated using a $\mathtt{softmax}$ function:
	\begin{equation}
		p(z|\mathbf{p}_t) = \frac{\exp(\mathbf{g}_z^\top \mathbf{p}_t + \mathbf{q}_z)}{\Sigma_{z^\prime\in \mathcal{A}(S, B)}\exp(\mathbf{g}_{z^\prime}^\top \mathbf{p}_t + \mathbf{q}_{z^\prime})}
		\label{eqn:2}
	\end{equation}
	where $\mathcal{A}(S, B)$ represents the set of valid actions given the current content in the \textit{stack} and \textit{buffer}.
	
	We apply the non-projective transition system originally introduced by~\newcite{nivre2009non} since most of the treebanks we consider in this study has a noticeable proportion of non-projective trees.
	In the \textsc{Swap}-based system, both the \textit{stack} and \textit{buffer} may contain tree fragments, so RecNN is applied both in S and B to obtain representations of each position.
	
	
	\subsection{Deep Multi-task Learning}
	\label{subsec:mtl}
	
	Multi-task learning (MTL) is the procedure of inductive transfer that improves learning for one task by using the information contained in the training signals of other related tasks.
	It does this by learning tasks in parallel while using a shared representation.
	A good overview, especially focusing on neural networks, can be found in~\newcite{caruana1997multitask}.
	
	\begin{figure}[t]
		\centering
		\includegraphics[width=80mm]{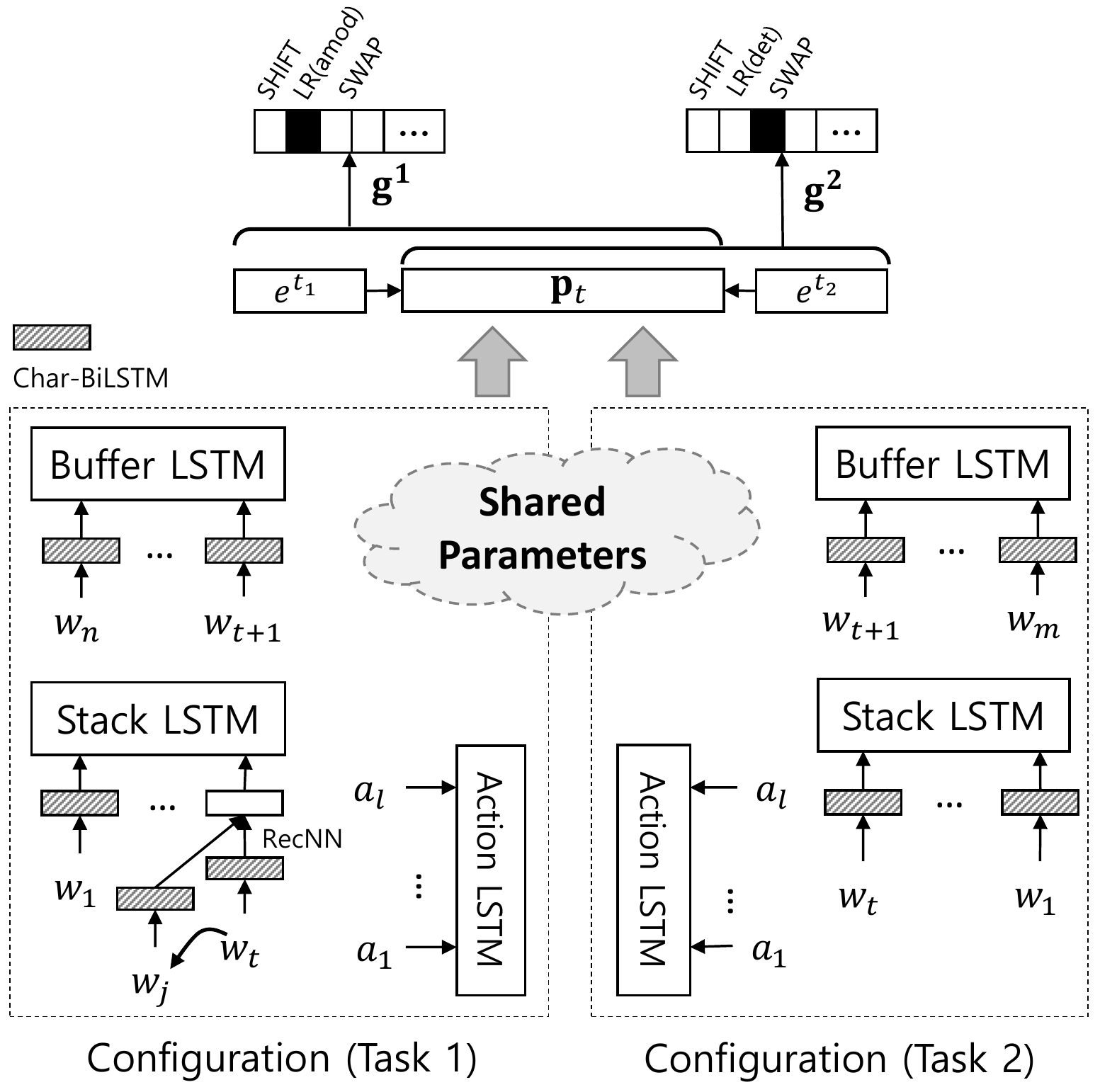}
		\caption{The architecture of deep multi-task learning.}\label{fig:multitask}
		\vspace{-0.5em}
	\end{figure}
	
	We illustrate our multi-task learning architecture in Figure~\ref{fig:multitask}.
	As discussed in previous sections, multiple treebanks, either multilingual or monolingual heterogeneous, contain knowledge that can be mutually beneficial.
	We consider the target treebank processing as the \textit{primary task}, and the source treebank as a \textit{related task}. 
	The two tasks are interacted through multi-level parameter sharing (Section~\ref{subsubsec:param-share}).
	Inspired by~\newcite{waleed2016unidep}, we introduce a task-specific vector $e^t$ (\textit{task embedding}) which is first combined with $\mathbf{s}_t, \mathbf{b}_t, \mathbf{a}_t$ to compute $\mathbf{p}_t$, and then further concatenated with $\mathbf{p}_t$ to compute the probability distribution of transition actions.
	Therefore, Eqn~\ref{eqn:1},~\ref{eqn:2} become:
	\begin{equation}
		\mathbf{p}_t = \mathtt{ReLU}(\mathbf{W}[\mathbf{s}_t; \mathbf{b}_t; \mathbf{a}_t; e^t] + \mathbf{d})
		\label{eqn:3}
	\end{equation}
	\begin{equation}
		p(z|\mathbf{p}_t) = \mathtt{softmax}(\mathbf{g}_z^\top [\mathbf{p}_t; e^t] + \mathbf{q}_z)
		\label{eqn:4}
	\end{equation}
		
	Each task uses the same formalism for optimization, and the joint cross-entropy is used as the objective function.
	The key of \textit{multi-task} learning is \textit{parameter sharing}, without which the correlation between tasks will not be exploited.
	Conventional multi-task learning models typically share a small proportion of parameters across tasks.
	For example, \newcite{collobert2008unified} only shares word embeddings, and \newcite{dong-EtAl:2015:ACL-IJCNLP2} shares the encoder of sequence-to-sequence models.
	In this work, we use more sophisticated parameter sharing strategies according to the linguistic similarities and differences between the tasks.
	
	\subsubsection{Parameter Sharing}
	\label{subsubsec:param-share}
	Deep neural networks automatically learn features for a specific task with hierarchical abstractions, which gives us the flexibility to control parameter sharing in different levels accordingly.
	
	In this study, different parameter sharing strategies are applied according to the source and target treebanks being used.
	We consider two different scenarios: MTL with multilingual universal treebanks as source (\textbf{\textsc{Multi-Univ}}) and MTL with monolingual heterogeneous treebanks as source (\textbf{\textsc{Mono-Hetero}}).
	Table~\ref{tbl:param-sharing-multilingual} presents our parameter sharing strategies for each setting.
	\begin{table}[t]
		\small
		\centering
		\begin{tabular}{l|l|l}
			\hline
			& \textsc{\textbf{Multi-Univ}} & \textsc{\textbf{Mono-Hetero}} \\
			\hline
			\bf Shared & \specialcell{LSTM(S) \\ LSTM(B) \\ RecNN \\ $W_A, W_S, W_B$ \\ $E_{pos}, E_{rel}, E_{act}$} & \specialcell{LSTM(S) \\ LSTM(B) \\ BiLSTM(chars) \\ RecNN \\ $W_A, W_S, W_B$ \\ $E_{pos}, E_{char}$ } \\
			\hline
			\bf \specialcell{Task-\\specific} & \specialcell{LSTM(A) \\ BiLSTM(chars) \\ $\mathbf{g}$ \\ $E_{char}, e^t$} & \specialcell{LSTM(A) \\ $\mathbf{g}$ \\ $E_{rel}, E_{act}, e^t$} \\
			\hline
		\end{tabular}
		\caption{Parameter sharing strategies for \textsc{\textbf{Multi-Univ}} and \textsc{\textbf{Mono-Hetero}}. LSTM(S) -- \textit{stack} LSTM; LSTM(B) -- \textit{buffer} LSTM; LSTM(A) -- \textit{action} LSTM; BiLSTM(chars) -- Char-BiLSTM; RecNN -- recursive NN modeling the subtrees; $W_A, W_S, W_B$ -- weights from A, S, B to the state ($\mathbf{p}_t$); $\mathbf{g}$ -- weights from the state to output layer; $E$ -- embeddings.}
		\label{tbl:param-sharing-multilingual}
		\vspace{-0.5em}
	\end{table}
	
	\textsc{\textbf{Multi-Univ}}.
	Multilingual universal treebanks are annotated with the same set of POS tags~\cite{petrov2011universal}, dependency relations, and thus share the same set of transition actions.
	However, the vocabularies (word, characters) are language-specific.
	Additionally, linguistic typologies such as the order of \textit{subject-verb-object} and \textit{adjective-noun} (Figure~\ref{fig:multilingual}) also varies across languages, which result in the divergence of inherent grammars of transition actions.
	Therefore, it makes sense to share the lookup tables (embeddings) of POS tags ($E_{pos}$), relations ($E_{rel}$) and actions ($E_{act}$), but separate the character embeddings ($E_{char}$) as well as the Char-BiLSTM (BiLSTM(chars)), and  also the LSTM modeling action sequence (LSTM(A))
	
	\textsc{\textbf{Mono-Hetero}}.
	Monolingual heterogeneous treebanks instead share the same lexical representations, but have different POS tags, structures and relations (Figure~\ref{fig:heterogeneous}) due to the different annotation schemes.
	Hence the transition actions set varies across treebanks.
	For simplicity reasons, we convert the language-specific POS tags in the heterogeneous treebanks into universal POS tags~\cite{petrov2011universal}.
	Consequently, $E_{char}$ and BiLSTM(chars), $E_{pos}$ are shared across tasks, but $E_{rel}$, $E_{act}$, LSTM(A) are separated.
	
	Besides, the LSTM parameters for modeling the \textit{stack} and \textit{buffer} (LSTM(S), LSTM(B)), the RecNN for modeling tree compositions, and the weights from S, B, A to the state $\mathbf{p}_t$ ($W_A, W_B, W_S$) are shared for both \textsc{Multi-Univ} and \textsc{Mono-Hetero}.
	As standard in multi-task learning, the weights at the output layer ($\mathbf{g}$) are \textit{task-specific} in both settings.

	\subsubsection{Learning}
	\label{subsubsec:learning}
	Training is achieved in a stochastic manner by looping over the tasks:
	
	\begin{enumerate}
		\vspace{-0.7em}
		\item Randomly select a task.
		\vspace{-0.7em}
		\item Select a sentence from the task, and generate instances for classification.
		\vspace{-0.7em}
		\item Update the corresponding parameters by back-propagation w.r.t. the instances.
		\vspace{-0.7em}
		\item Go to 1.
		\vspace{-0.7em}
	\end{enumerate}
	
	We adopt the development data of the target treebank (primary task) for early-stopping.
	
	\section{Experiments}
	We first describe the data and settings in our experiments, then the results and analysis.
	
	\subsection{Data and Settings}
	We conduct experiments on UDT v2.0\footnote{\url{github.com/ryanmcd/uni-dep-tb}} and the CoNLL-X shared task data.
	For monolingual heterogeneous source, we also experiment on CTB5 using CDT as the source treebank, to compare with the previous work of~\newcite{li-liu-che:2012:ACL2012}.
	Statistics of the datasets are summarized in Table~\ref{tbl:udt-stat}.
	We investigate the following experiment settings:
	\begin{itemize}
		\item \textsc{Multilingual} (\textsc{Univ}$\rightarrow$\textsc{Univ}).
		In this setting, we study the integration of multilingual universal treebanks.
		Experiments are conducted using the \textsc{UDT} dataset.
		Specifically, we consider \textsc{DE}, \textsc{ES}, \textsc{FR}, \textsc{PT}, \textsc{IT} and \textsc{SV} treebanks as target treebanks, and the \textsc{EN} treebank as the common source treebank.
		\item \textsc{Monolingual} (\textsc{Conll}$\leftrightarrow$\textsc{Univ}).
		Here we study the integration of monolingual heterogeneous treebanks.
		The \textsc{CoNLL-X} corporas (\textsc{DE}, \textsc{ES}, \textsc{PT}, \textsc{SV}) and the \textsc{UDT} treebank of corresponding languages are used as source and target treebanks mutually.
		\item \textsc{Monolingual} (\textsc{CDT}$\rightarrow$\textsc{CTB5}).
		We follow the same settings of~\newcite{li-liu-che:2012:ACL2012}, and consider two scenarios using automatic POS tags and gold-standard POS tags respectively.
	\end{itemize}
	
	We use the widely-adopted unlabeled attachment score (UAS) and labeled attachment score (LAS) for evaluation.
	
	\begin{table}[t]
		\centering
		\small
		\begin{tabular}{c|C{0.8cm}C{0.8cm}C{0.7cm}|C{0.8cm}C{0.8cm}C{0.7cm}}
			\hline
			& Train & Dev & Test & Train & Dev & Test \\
			\hline
			& \multicolumn{3}{c|}{\textsc{UDT}} & \multicolumn{3}{c}{\textsc{CoNLL-X}} \\
			\cdashline{2-7}
			\textsc{EN} & 39,832 & 1,700 & 2,416 & -- & -- & -- \\
			\textsc{DE} & 14,118 & 800 & 1,000 & 35,295 & 3,921 & 357 \\
			\textsc{ES} & 14,138 & 1,569 & 300 & 2,976 & 330 & 206 \\
			\textsc{FR} & 14,511 & 1,611 & 300 & -- & -- & -- \\
			\textsc{PT} & 9,600 & 1,200 & 1,198 & 8,164 & 907 & 288 \\
			\textsc{IT} & 6,389 & 400 & 400 & -- & -- & -- \\
			\textsc{SV} & 4,447 & 493 & 1,219 & 9,938 & 1,104 & 389 \\
			\hline
			& \multicolumn{3}{c|}{\textsc{CDT}} & \multicolumn{3}{c}{\textsc{CTB5}} \\
			\cdashline{2-7}
			\textsc{ZH} & 55,500 & 1,500 & 3,000 & 16,091 & 803 & 1,910 \\
			\hline
		\end{tabular}
		\caption{Statics of UDT v2.0 and CoNLL-X treebanks (with languages presented in UDT v2.0).}
		\label{tbl:udt-stat}
		\vspace{-0.5em}
	\end{table}
	
	\subsection{Baseline Systems}
	\label{subsec:baseline}
	We compare our approach with the following baseline systems.
	\begin{itemize}
		\item Monolingual supervised training (\textsc{Sup}).
		Models are trained only on the target treebank, with the \textsc{LSTM}-based parser.
		\item Cascaded training (\textsc{Cas}).
		This system has two stages.
		First, models are trained using the source treebank.
		Then the parameters are used to initialize the neural network for training target parsers.
		Similar approach was studied in~\newcite{duong-EtAl:2015:ACL-IJCNLP} and \newcite{guo2016multisource} for low-resource parsing.
	\end{itemize}
	
	For \textsc{Multilingual} (\textsc{Univ}$\rightarrow$\textsc{Univ}), we also compare with the \textit{shallow multi-task learning} (\textsc{SMTL}) system, as described in Section~\ref{sec:related-work}, which is representative of the approach of~\newcite{duong-EtAl:2015:EMNLP} and \newcite{waleed2016unidep}.
	In \textsc{SMTL} all the parameters are shared except the character embeddings ($E_{char}$), and task embeddings ($e^t$) are not used.
	Unlike~\newcite{duong-EtAl:2015:EMNLP} and ~\newcite{waleed2016unidep}, we don't use external resources such as cross-lingual word clusters, embeddings and dictionaries which is beyond the scope of this work.
	

	\begin{table*}[t]
		\centering
		\small
		\begin{tabular}{c|cc|cc|cc|cc}
			\hline
			& \multicolumn{8}{c}{\textsc{Multilingual} (\textsc{Univ} $\rightarrow$ \textsc{Univ})} \\
			\cline{2-9}
			& \multicolumn{2}{c|}{\textsc{Sup}} & \multicolumn{2}{c|}{$\textsc{Cas}_{EN}$} & \multicolumn{2}{c|}{$\textsc{SMTL}_{EN}$} & \multicolumn{2}{c}{$\textsc{MTL}_{EN}$} \\ 
			& UAS & LAS & UAS & LAS & UAS & LAS & UAS & LAS \\ 
			\hline
			\textsc{DE} & 84.24 & 78.40 & 84.24 & 78.65 & 84.37 & 79.07 & \bf 84.93 & \bf 79.34 \\ 
			\textsc{ES} & 85.31 & 81.23 & 85.42 & 81.42 & 85.78 & 81.54 & \bf 86.78 & \bf 82.92 \\ 
			\textsc{FR} & 85.55 & 81.13 & 84.57 & 80.14 & 86.13 & 81.77 & \bf 86.44 & \bf 82.01 \\ 
			\textsc{PT} & 88.40 & 86.54 & 88.88 & 87.07 & 89.08 & 87.24 & \bf 89.24 & \bf 87.50 \\ 
			\textsc{IT} & 86.53 & 83.72 & 86.58 & 83.67 & 86.53 & 83.64 & \bf 87.26 & \bf 84.27 \\ 
			\textsc{SV} & 84.91 & 79.88 & 86.43 & 81.92 & \bf 86.79 & \bf 82.31 & 85.98 & 81.35 \\ 
			\cdashline{1-9}
			\it \textsc{Avg} &\it 85.82 &\it 81.82 &\it 86.02 &\it 82.15 &\it 86.45 &\it 82.60 &\bf \textit{86.77} & \bf \textit{82.90} \\ 
			\hline
		\end{tabular}
		\caption{Parsing accuracies of \textsc{Multilingual} (\textsc{Univ}$\rightarrow$\textsc{Univ}). Significance tests with MaltEval yield p-values $<$ 0.01 for ($\textsc{MTL}$ \textit{vs.} \textsc{Sup}) on all languages.}
		\label{tbl:multi-uni-uni}
		\vspace{-0.5em}
	\end{table*}
	
	\subsection{Results}
	In this section, we present empirical evaluations under different settings.

	\subsubsection{Multilingual Universal Source Treebanks}
	Table~\ref{tbl:multi-uni-uni} shows the results under the \textsc{Multilingual} (\textsc{Univ}$\rightarrow$\textsc{Univ}) setting.
	\textsc{Cas} yields slightly better performance than \textsc{Sup}, especially for SV (+1.52\% UAS and +2.04\% LAS), indicating that \textit{pre-training} with EN training data indeed provides a better initialization of the parameters for cascaded training.
	\textsc{SMTL} in turn outperforms \textsc{Cas} overall (comparable for \textsc{IT}), which implies that training two treebanks jointly helps even with an unique model.
	
	Furthermore, with appropriate parameter sharing, our deep multi-task learning approach (\textsc{MTL}) outperforms \textsc{Sup} overall and achieves the best performances in five out of six languages.
	An exception is Swedish.
	As we can see, both \textsc{Cas} and \textsc{SMTL} outperforms \textsc{MTL} by a significant margin for \textsc{SV}.	
	The underlying reasons we suggest are two-fold.
	\begin{enumerate}
		\item \textsc{SV} morphology is similar to \textsc{EN} with less inflections, encouraging the morphology-related parameters like BiLSTM(chars) to be shared.
		\vspace{-0.5em}
		\item \textsc{SV} has a much smaller treebank compared with \textsc{EN} (1:9).
		Intuitively, \textsc{SMTL} and \textsc{Cas} work better in low resource setting.
	\end{enumerate}
	
	To verify the first issue, we conduct tests on \textsc{SMTL} without sharing Char-BiLSTMs.
	As shown in Table~\ref{tbl:effect-smtl-sv}, the performance of \textsc{SMTL} decreases significantly (-0.73 in UAS).
	This observation also indicates that \textsc{MTL} has the potential to reach higher performances through \textit{language-specific tuning} of parameter sharing strategies.
	\begin{table}[t]
		\centering
		\small
		\begin{tabular}{c|l|cc}
			\hline
			\multicolumn{2}{c|}{} & UAS & LAS \\
			\hline
			\multirow{2}{*}{\textsc{SV}} & \textsc{SMTL} & 86.79 & 82.31 \\
			& ~~~-- \textit{shared}-BiLSTM(chars) & 86.06 & 81.50 \\
			\hline
		\end{tabular}
		\caption{\textsc{SMTL} for Swedish without sharing BiLSTM(chars).}
		\label{tbl:effect-smtl-sv}
		\vspace{-0.5em}
	\end{table}
	
	To verify the second issue, we consider a low resource setup following~\newcite{duong-EtAl:2015:EMNLP}, where the target language has a small treebank (3K tokens).
	We train our models on identical sampled dataset shared by~\newcite{duong-EtAl:2015:EMNLP} on \textsc{DE}, \textsc{ES} and \textsc{FR}.
	As we can find in Table~\ref{tbl:low-resource}, while all the models outperform \textsc{Sup}, both \textsc{Cas} and \textsc{SMTL} work better than \textsc{MTL}, which confirms our assumption.
	Although not the primary focus of this work, we find that \textsc{SMTL} and \textsc{MTL} can be significantly improved in low resource setting through weighted sampling of tasks during training.
	Specifically, in the training procedure (Section \ref{subsubsec:learning}), we sample from the source language (\textsc{EN}) which has a much richer treebank with larger probability of 0.9, while sample from the target language with probability of 0.1.
	In this way, the two tasks are encouraged to converge at a similar rate.
	As shown in Table~\ref{tbl:low-resource}, both \textsc{SMTL} and \textsc{MTL} benefit from weighted task sampling.
	
	\begin{table}[t]
		\centering
		\small
		\begin{tabular}{l|c|c|c}
		\hline
			& \textsc{DE} & \textsc{ES} & \textsc{FR} \\
		\hline
		\textsc{Sup} & 58.93 & 61.99 & 60.45 \\
		\textsc{Cas} & 64.08 & \bf 70.45 & \bf 68.72 \\
		\textsc{SMTL} & 63.57 & 69.01 & 65.04 \\
		\it ~~~+ weighted sampling & \it 63.50 & \it 70.17 & \it 68.52 \\
		\textsc{MTL} & 62.43 & 66.67 & 64.23 \\
		\it ~~~+ weighted sampling & \it \textbf{64.22} & \it 68.42 & \it 66.67 \\
		\hline
		Duong et al. & 61.2 & 69.1 & 65.3 \\
		Duong et al. + Dict & 61.8 & 70.5 & 67.2 \\
		\hline
		\end{tabular}
		\caption{Low resource setup (3K tokens), evaluated with LAS.}
		\label{tbl:low-resource}
		\vspace{-1.0em}
	\end{table}
	
	
	\subsubsection{Monolingual Hetero. Source Treebanks}
		
	\begin{table}[t]
		\centering
		\small
		\begin{tabular}{c|C{0.75cm}C{0.75cm}|C{0.75cm}C{0.75cm}|C{0.75cm}C{0.75cm}}
			\hline
			& \multicolumn{2}{c|}{\textsc{Sup}} & \multicolumn{2}{c|}{$\textsc{Cas}$} & \multicolumn{2}{c}{$\textsc{MTL}$} \\
			& UAS & LAS & UAS & LAS & UAS & LAS \\
			\hline
			& \multicolumn{6}{c}{\textsc{Monolingual} (\textsc{Conll}$\rightarrow$\textsc{Univ})} \\
			\cline{2-7}
			\textsc{DE} & 84.24 & 78.40 & 85.02 & 80.05 & \bf 85.73 & \bf 80.64 \\
			\textsc{ES} & 85.31 & 81.23 & \bf 85.90 & \bf 81.73 & 85.80 & 81.45 \\
			\textsc{PT} & 88.40 & 86.54 & 89.12 & 87.32 & \bf 89.40 & \bf 87.60 \\
			\textsc{SV} & 84.91 & 79.88 & 87.17 & 82.83 & \bf 87.27 & \bf 83.52 \\
			\textsc{SV$^*$} & 82.61 & 77.42 & \bf 85.39 & 80.60 & 85.29 & \bf 81.22 \\
			\cdashline{1-7}
			\it \textsc{Avg}&\it 85.14 &\it 80.90 &\it 86.35 &\it 82.43 &\bf \textit{86.56} &\bf \textit{82.73} \\
			\hline
			& \multicolumn{6}{c}{\textsc{Monolingual} (\textsc{Univ}$\rightarrow$\textsc{Conll})} \\
			\cline{2-7}
			\textsc{DE} & 89.06 & 86.48 & 89.64 & 86.66 & \bf 89.98 & \bf 87.50 \\
			\textsc{ES} & 85.41 & 80.50 & \bf 86.46 & 81.37 & 86.07 & \bf 81.41 \\
			\textsc{PT} & \bf 90.16 & \bf 85.53 & 89.50 & 85.03 & 89.98 & 85.23 \\
			\textsc{SV} & 88.49 & 81.98 & 89.07 & 82.91 & \bf 91.60 & \bf 85.22 \\
			\textsc{SV$^*$} & 79.61 & 72.71 & 82.91 & 74.96 & \bf 84.86 & \bf 77.36 \\
			\cdashline{1-7}
			\it \textsc{Avg}&\it 86.06 &\it 81.31 &\it 87.13 &\it 82.01 &\bf \textit{87.72} & \bf \textit{82.88} \\
			\hline
		\end{tabular}
		\caption{\textsc{Monolingual} (\textsc{Conll}$\leftrightarrow$\textsc{Univ}) performance. \textsc{SV$^*$} is used for computing the \textsc{Avg} values.}
		\label{tbl:mono-conllx-uni}
		\vspace{-0.5em}
	\end{table}
	
		\begin{table*}[t]
			\centering
			\small
			\begin{tabular}{ll|ccc|ccc}
				\hline
				& & \multicolumn{3}{c|}{Auto-POS} & \multicolumn{3}{c}{Gold-POS} \\
				\multicolumn{2}{c|}{} & \textsc{Sup} & \textsc{Cas} & \textsc{MTL} & \textsc{Sup} & \textsc{Cas} & \textsc{MTL} \\
				\multirow{2}{*}{\textsc{Ours}} & UAS & 79.34 & 80.25 (+0.91) & \bf 81.13 (+1.79) & 85.25 & 86.29 (+1.04) & \bf 86.69 (+1.44) \\
				& LAS & 76.23 & 77.26 (+1.03) & \bf 78.24 (+2.01) & 83.59 & 84.72 (+1.13) & \bf 85.18 (+1.59) \\
				\hline
				\multicolumn{2}{c|}{} & \textsc{Sup} & \multicolumn{2}{c|}{with QG} & \textsc{Sup} & \multicolumn{2}{c}{with QG} \\
				\textsc{Li12-O2} & \multirow{2}{*}{UAS} & 79.67 & \multicolumn{2}{c|}{81.04 (+1.37)} & 86.13 & \multicolumn{2}{c}{86.44 (+0.31)} \\
				\textsc{Li12-O2sib} & & 79.25 & \multicolumn{2}{c|}{80.45 (+1.20)} & 85.63 & \multicolumn{2}{c}{86.17 (+0.54)} \\
				\hline
			\end{tabular}
			\caption{Parsing accuracy comparisons of \textsc{Monolingual} (\textsc{CDT}$\rightarrow$\textsc{CTB5}). \textsc{Li12-O2} use the O2 graph-based parser with both sibling and grandparent structures, while \textsc{Li12-O2sib} only use the sibling parts (Li et al., 2012).}
			\label{tbl:mono-cdt-ctb5}
			\vspace{-0.5em}
		\end{table*}

	Table~\ref{tbl:mono-conllx-uni} shows the results of \textsc{Monolingual} (\textsc{Conll}$\leftrightarrow$\textsc{Univ}).
	Overall \textsc{MTL} systems outperforms the supervised baselines by significant margins in both conditions, showing the mutual benefits of \textsc{UDT} and \textsc{CoNLL-X} treebanks.\footnote{An exception is \textsc{PT} in \textsc{Monolingual} (\textsc{Univ}$\rightarrow$\textsc{Conll}), in which both \textsc{Cas} and \textsc{MTL} get slightly degradation in performance. This may be due to the low quality of the PT universal treebank caused by the automatic construction process. We discussed and verified this with the author of UDT v2.0.}
	
	
	
	In addition, among the four languages here, the \textsc{SV} universal treebank is mainly converted from the Talbanken part of the Swedish bank~\cite{nivre2007bootstrapping}, thus has a large overlap with the \textsc{CoNLL-X} Swedish treebank.
	In fact, we find a large proportion of the \textsc{SV} test data in UDT/\textsc{CoNLL-X} appears in \textsc{CoNLL-X}/UDT \textsc{SV} training data.
	Typically we expect fully \textit{unseen} data for testing, so we further separate the SV testing data into two parts: \textsc{In-Src} and \textsc{Out-Src} including sentences that appear in the source treebank or not, respectively.
	Statistics are shown below.
	\begin{table}[h]
		\centering
		\small
		\vspace{-0.5em}
		\begin{tabular}{l|cc}
			& \textsc{Conll}$\rightarrow$\textsc{Univ} & \textsc{Univ}$\rightarrow$\textsc{Conll} \\
			\hline
			\textsc{In-Src} & 875 & 352 \\
			\textsc{Out-Src} & 344 & 37 \\
		\end{tabular}
		\vspace{-0.5em}
	\end{table}

	The \textsc{SV$^*$} row in Table~\ref{tbl:mono-conllx-uni} presents the \textsc{Out-Src} results of SV, which shows consistent improvements.

	To show the merit of our approach against previous approaches, we further conduct experiments on CTB5 using CDT as heterogeneous source treebank (Table~\ref{tbl:udt-stat}). 
	For CTB5, we follow~\cite{li-liu-che:2012:ACL2012} and consider two scenarios which use automatic POS tags and gold-standard POS tags respectively.
	To compare with their results, we run \textsc{Sup}, \textsc{Cas} and \textsc{MTL} on CTB5.
	Table~\ref{tbl:mono-cdt-ctb5} presents the results.
	The indirect comparison indicates that our approach can achieve larger improvement than their method in both scenarios.
	Beside the empirical comparison, our method has the additional advantages in its scalability to multi-typed source treebanks without the painful human efforts of feature design.
	
	
	\subsection{Remarks}
	Overall, our approach obtains substantial gains over supervised baselines with either multilingual universal treebanks or monolingual heterogeneous treebanks as source.
	With multilingual source treebanks, our model has the potential to improve even further via language-specific tuning.
	While not the primary focus of this study, in low resource setting, we show that more emphasize may be put on the source treebanks through weighted task sampling.
	
	

	\section{Conclusion}
	This paper propose an universal framework based on deep multi-task learning that can integrate arbitrary-typed source treebanks to enhance the parsing models on target treebanks.
	We study two scenarios, respectively using multilingual universal source treebanks and monolingual heterogeneous source treebanks, and design effective parameter sharing strategies for each scenario.
	
	We conduct extensive experiments on several benchmark treebanks in various languages. 
	Results demonstrate that our approach significantly improves over baseline systems under various experiment setting.
	Furthermore, our framework can flexibly incorporate richer treebanks and more related tasks, which we leave to future exploration.
	
	\section*{Acknowledgments}
	We thank Ryan McDonald for fruitful discussions, and thank Dr. Zhenghua Li for sharing the processed CTB and CDT dataset. This work was supported by the National Key Basic Research Program of China via grant 2014CB340503 and the National Natural Science Foundation of China (NSFC) via grant 61133012 and 61370164.
	
	%

	\bibliographystyle{emnlp2016}

\begin{thebibliography}{}
		
		\bibitem[\protect\citename{Ammar \bgroup et al.\egroup }2016]{waleed2016unidep}
		Waleed Ammar, George Mulcaire, Miguel Ballesteros, Chris Dyer, and Noah~A
		Smith.
		\newblock 2016.
		\newblock One parser, many languages.
		\newblock {\em arXiv preprint arXiv:1602.01595}.
		
		\bibitem[\protect\citename{Ando and Zhang}2005]{Ando:2005:FLP:1046920.1194905}
		Rie~Kubota Ando and Tong Zhang.
		\newblock 2005.
		\newblock A framework for learning predictive structures from multiple tasks
		and unlabeled data.
		\newblock {\em JMLR}, 6:1817--1853, December.
		
		\bibitem[\protect\citename{Andor \bgroup et al.\egroup
		}2016]{andor2016globally}
		Daniel Andor, Chris Alberti, David Weiss, Aliaksei Severyn, Alessandro Presta,
		Kuzman Ganchev, Slav Petrov, and Michael Collins.
		\newblock 2016.
		\newblock Globally normalized transition-based neural networks.
		\newblock {\em arXiv preprint arXiv:1603.06042}.
		
		\bibitem[\protect\citename{Ballesteros \bgroup et al.\egroup
		}2015]{ballesteros-dyer-smith:2015:EMNLP}
		Miguel Ballesteros, Chris Dyer, and Noah~A. Smith.
		\newblock 2015.
		\newblock Improved transition-based parsing by modeling characters instead of
		words with lstms.
		\newblock In {\em Proc. of the 2015 Conference on EMNLP}, pages 349--359,
		September.
		
		\bibitem[\protect\citename{Bohnet and
			Nivre}2012]{bohnet-nivre:2012:EMNLP-CoNLL}
		Bernd Bohnet and Joakim Nivre.
		\newblock 2012.
		\newblock A transition-based system for joint part-of-speech tagging and
		labeled non-projective dependency parsing.
		\newblock In {\em Proc. of the 2012 Joint Conference on EMNLP and CoNLL}, pages
		1455--1465, July.
		
		\bibitem[\protect\citename{Buchholz and Marsi}2006]{buchholz2006conll}
		Sabine Buchholz and Erwin Marsi.
		\newblock 2006.
		\newblock Conll-x shared task on multilingual dependency parsing.
		\newblock In {\em Proc. of the Tenth Conference on Computational Natural
			Language Learning (CoNLL-X)}, pages 149--164, June.
		
		\bibitem[\protect\citename{Burkett and Klein}2008]{burkett-klein:2008:EMNLP}
		David Burkett and Dan Klein.
		\newblock 2008.
		\newblock Two languages are better than one (for syntactic parsing).
		\newblock In {\em Proc. of the 2008 Conference on EMNLP}, pages 877--886,
		October.
		
		\bibitem[\protect\citename{Caruana}1997]{caruana1997multitask}
		Rich Caruana.
		\newblock 1997.
		\newblock Multitask learning.
		\newblock {\em Machine learning}, 28(1):41--75.
		
		\bibitem[\protect\citename{Chen and Manning}2014]{chen2014fast}
		Danqi Chen and Christopher Manning.
		\newblock 2014.
		\newblock A fast and accurate dependency parser using neural networks.
		\newblock In {\em Proc. of the 2014 Conference on Empirical Methods in Natural
			Language Processing (EMNLP)}, pages 740--750, October.
		
		\bibitem[\protect\citename{Chen \bgroup et al.\egroup
		}2010]{chen-kazama-torisawa:2010:ACL}
		Wenliang Chen, Jun'ichi Kazama, and Kentaro Torisawa.
		\newblock 2010.
		\newblock Bitext dependency parsing with bilingual subtree constraints.
		\newblock In {\em Proc. of the 48th ACL}, pages 21--29, July.
		
		\bibitem[\protect\citename{Collobert and Weston}2008]{collobert2008unified}
		Ronan Collobert and Jason Weston.
		\newblock 2008.
		\newblock A unified architecture for natural language processing: Deep neural
		networks with multitask learning.
		\newblock In {\em Proc. of the 25th ICML}, pages 160--167.
		
		\bibitem[\protect\citename{Dong \bgroup et al.\egroup
		}2015]{dong-EtAl:2015:ACL-IJCNLP2}
		Daxiang Dong, Hua Wu, Wei He, Dianhai Yu, and Haifeng Wang.
		\newblock 2015.
		\newblock Multi-task learning for multiple language translation.
		\newblock In {\em Proc. of the 53rd ACL and the 7th IJCNLP (Volume 1: Long
			Papers)}, pages 1723--1732, July.
		
		\bibitem[\protect\citename{Duong \bgroup et al.\egroup
		}2015a]{duong-EtAl:2015:ACL-IJCNLP}
		Long Duong, Trevor Cohn, Steven Bird, and Paul Cook.
		\newblock 2015a.
		\newblock Low resource dependency parsing: Cross-lingual parameter sharing in a
		neural network parser.
		\newblock In {\em Proc. of the 53rd ACL and the 7th IJCNLP (Volume 2: Short
			Papers)}, pages 845--850, July.
		
		\bibitem[\protect\citename{Duong \bgroup et al.\egroup
		}2015b]{duong-EtAl:2015:EMNLP}
		Long Duong, Trevor Cohn, Steven Bird, and Paul Cook.
		\newblock 2015b.
		\newblock A neural network model for low-resource universal dependency parsing.
		\newblock In {\em Proc. of the 2015 Conference on EMNLP}, pages 339--348,
		September.
		
		\bibitem[\protect\citename{Dyer \bgroup et al.\egroup
		}2015]{dyer-EtAl:2015:ACL-IJCNLP}
		Chris Dyer, Miguel Ballesteros, Wang Ling, Austin Matthews, and Noah~A. Smith.
		\newblock 2015.
		\newblock Transition-based dependency parsing with stack long short-term
		memory.
		\newblock In {\em Proc. of the 53rd ACL and the 7th IJCNLP (Volume 1: Long
			Papers)}, pages 334--343, July.
		
		\bibitem[\protect\citename{Guo \bgroup et al.\egroup
		}2015]{guo-EtAl:2015:ACL-IJCNLP2}
		Jiang Guo, Wanxiang Che, David Yarowsky, Haifeng Wang, and Ting Liu.
		\newblock 2015.
		\newblock Cross-lingual dependency parsing based on distributed
		representations.
		\newblock In {\em Proc. of the 53rd ACL and the 7th IJCNLP (Volume 1: Long
			Papers)}, pages 1234--1244, July.
		
		\bibitem[\protect\citename{Guo \bgroup et al.\egroup }2016]{guo2016multisource}
		Jiang Guo, Wanxiang Che, David Yarowsky, Haifeng Wang, and Ting Liu.
		\newblock 2016.
		\newblock A representation learning framework for multi-source transfer
		parsing.
		\newblock In {\em Proc. of the Thirtieth AAAI Conference on Artificial
			Intelligence (AAAI)}, February.
		
		\bibitem[\protect\citename{Hatori \bgroup et al.\egroup
		}2012]{hatori-EtAl:2012:ACL2012}
		Jun Hatori, Takuya Matsuzaki, Yusuke Miyao, and Jun'ichi Tsujii.
		\newblock 2012.
		\newblock Incremental joint approach to word segmentation, pos tagging, and
		dependency parsing in chinese.
		\newblock In {\em Proc. of the 50th ACL (Volume 1: Long Papers)}, pages
		1045--1053, July.
		
		\bibitem[\protect\citename{Henderson and
			Titov}2010]{Henderson:2010:ISB:1756006.1953044}
		James Henderson and Ivan Titov.
		\newblock 2010.
		\newblock Incremental sigmoid belief networks for grammar learning.
		\newblock {\em J. Mach. Learn. Res.}, 11:3541--3570, December.
		
		\bibitem[\protect\citename{Henderson \bgroup et al.\egroup
		}2013]{henderson2013multilingual}
		James Henderson, Paola Merlo, Ivan Titov, and Gabriele Musillo.
		\newblock 2013.
		\newblock Multilingual joint parsing of syntactic and semantic dependencies
		with a latent variable model.
		\newblock {\em Computational Linguistics}, 39(4):949--998.
		
		\bibitem[\protect\citename{Huang \bgroup et al.\egroup
		}2009]{huang-jiang-liu:2009:EMNLP}
		Liang Huang, Wenbin Jiang, and Qun Liu.
		\newblock 2009.
		\newblock Bilingually-constrained (monolingual) shift-reduce parsing.
		\newblock In {\em Proc. of the 2009 Conference on EMNLP}, pages 1222--1231,
		August.
		
		\bibitem[\protect\citename{Hwa \bgroup et al.\egroup
		}2005]{hwa2005bootstrapping}
		Rebecca Hwa, Philip Resnik, Amy Weinberg, Clara Cabezas, and Okan Kolak.
		\newblock 2005.
		\newblock Bootstrapping parsers via syntactic projection across parallel texts.
		\newblock {\em Natural language engineering}, 11(03):311--325.
		
		\bibitem[\protect\citename{Johansson}2013]{johansson:2013:NAACL-HLT}
		Richard Johansson.
		\newblock 2013.
		\newblock Training parsers on incompatible treebanks.
		\newblock In {\em Proc. of NAACL: HLT}, pages 127--137, June.
		
		\bibitem[\protect\citename{Li \bgroup et al.\egroup }2011]{li-EtAl:2011:EMNLP3}
		Zhenghua Li, Min Zhang, Wanxiang Che, Ting Liu, Wenliang Chen, and Haizhou Li.
		\newblock 2011.
		\newblock Joint models for chinese pos tagging and dependency parsing.
		\newblock In {\em Proc. of the 2011 Conference on EMNLP}, pages 1180--1191,
		July.
		
		\bibitem[\protect\citename{Li \bgroup et al.\egroup
		}2012]{li-liu-che:2012:ACL2012}
		Zhenghua Li, Ting Liu, and Wanxiang Che.
		\newblock 2012.
		\newblock Exploiting multiple treebanks for parsing with quasi-synchronous
		grammars.
		\newblock In {\em Proc. of the 50th ACL (Volume 1: Long Papers)}, pages
		675--684, July.
		
		\bibitem[\protect\citename{Luong \bgroup et al.\egroup
		}2015]{luong2015multitask}
		Minh{-}Thang Luong, Quoc~V. Le, Ilya Sutskever, Oriol Vinyals, and Lukasz
		Kaiser.
		\newblock 2015.
		\newblock Multi-task sequence to sequence learning.
		\newblock {\em CoRR}, abs/1511.06114.
		
		\bibitem[\protect\citename{Marcus \bgroup et al.\egroup
		}1993]{marcus1993building}
		Mitchell~P Marcus, Mary~Ann Marcinkiewicz, and Beatrice Santorini.
		\newblock 1993.
		\newblock Building a large annotated corpus of english: The penn treebank.
		\newblock {\em Computational linguistics}, 19(2):313--330.
		
		\bibitem[\protect\citename{McDonald \bgroup et al.\egroup
		}2011]{mcdonald2011multi}
		Ryan McDonald, Slav Petrov, and Keith Hall.
		\newblock 2011.
		\newblock Multi-source transfer of delexicalized dependency parsers.
		\newblock In {\em Proc. of the 2011 Conference on EMNLP}, pages 62--72, July.
		
		\bibitem[\protect\citename{McDonald \bgroup et al.\egroup
		}2013]{mcdonald2013universal}
		Ryan McDonald, Joakim Nivre, Yvonne Quirmbach-Brundage, Yoav Goldberg, Dipanjan
		Das, Kuzman Ganchev, Keith Hall, Slav Petrov, Hao Zhang, Oscar
		T\"{a}ckstr\"{o}m, Claudia Bedini, N\'{u}ria Bertomeu~Castell\'{o}, and
		Jungmee Lee.
		\newblock 2013.
		\newblock Universal dependency annotation for multilingual parsing.
		\newblock In {\em Proc. of the 51st ACL (Volume 2: Short Papers)}, pages
		92--97, August.
		
		\bibitem[\protect\citename{Niu \bgroup et al.\egroup
		}2009]{niu-wang-wu:2009:ACLIJCNLP}
		Zheng-Yu Niu, Haifeng Wang, and Hua Wu.
		\newblock 2009.
		\newblock Exploiting heterogeneous treebanks for parsing.
		\newblock In {\em Proc. of the Joint Conference of the 47th ACL and the 4th
			IJCNLP of the AFNLP}, pages 46--54, August.
		
		\bibitem[\protect\citename{Nivre and
			McDonald}2008]{nivre-mcdonald:2008:ACLMain}
		Joakim Nivre and Ryan McDonald.
		\newblock 2008.
		\newblock Integrating graph-based and transition-based dependency parsers.
		\newblock In {\em Proc. of ACL-08: HLT}, pages 950--958, June.
		
		\bibitem[\protect\citename{Nivre and Megyesi}2007]{nivre2007bootstrapping}
		Joakim Nivre and Beata Megyesi.
		\newblock 2007.
		\newblock Bootstrapping a swedish treebank using cross-corpus harmonization and
		annotation projection.
		\newblock In {\em Proc. of the 6th International Workshop on Treebanks and
			Linguistic Theories}, pages 97--102.
		
		\bibitem[\protect\citename{Nivre}2009]{nivre2009non}
		Joakim Nivre.
		\newblock 2009.
		\newblock Non-projective dependency parsing in expected linear time.
		\newblock In {\em Proc. of the Joint Conference of the 47th Annual Meeting of
			the ACL and the 4th IJCNLP of the AFNLP}, pages 351--359, August.
		
		\bibitem[\protect\citename{Petrov \bgroup et al.\egroup
		}2012]{petrov2011universal}
		Slav Petrov, Dipanjan Das, and Ryan McDonald.
		\newblock 2012.
		\newblock A universal part-of-speech tagset.
		\newblock In {\em Proc. of the Eighth International Conference on Language
			Resources and Evaluation (LREC-2012)}, pages 2089--2096, May.
		
		\bibitem[\protect\citename{Rasooli and
			Collins}2015]{rasooli-collins:2015:EMNLP}
		Mohammad~Sadegh Rasooli and Michael Collins.
		\newblock 2015.
		\newblock Density-driven cross-lingual transfer of dependency parsers.
		\newblock In {\em Proc. of the 2015 Conference on EMNLP}, pages 328--338,
		September.
		
		\bibitem[\protect\citename{Smith and Eisner}2009]{smith-eisner:2009:EMNLP}
		David~A. Smith and Jason Eisner.
		\newblock 2009.
		\newblock Parser adaptation and projection with quasi-synchronous grammar
		features.
		\newblock In {\em Proc. of the 2009 Conference on EMNLP}, pages 822--831,
		August.
		
		\bibitem[\protect\citename{T\"{a}ckstr\"{o}m \bgroup et al.\egroup
		}2012]{tackstrom2012cross}
		Oscar T\"{a}ckstr\"{o}m, Ryan McDonald, and Jakob Uszkoreit.
		\newblock 2012.
		\newblock Cross-lingual word clusters for direct transfer of linguistic
		structure.
		\newblock In {\em Proc. of NAACL: HLT}, pages 477--487, June.
		
		\bibitem[\protect\citename{Tiedemann}2014]{tiedemann2014rediscovering}
		J\"{o}rg Tiedemann.
		\newblock 2014.
		\newblock Rediscovering annotation projection for cross-lingual parser
		induction.
		\newblock In {\em Proc. of COLING 2014}, pages 1854--1864, August.
		
		\bibitem[\protect\citename{Titov and
			Henderson}2007]{titov-henderson:2007:EMNLP-CoNLL2007}
		Ivan Titov and James Henderson.
		\newblock 2007.
		\newblock Fast and robust multilingual dependency parsing with a generative
		latent variable model.
		\newblock In {\em Proc. of the CoNLL Shared Task Session of EMNLP-CoNLL 2007},
		pages 947--951, June.
		
		\bibitem[\protect\citename{Torres~Martins \bgroup et al.\egroup
		}2008]{torresmartins-EtAl:2008:EMNLP}
		Andr\'{e}~Filipe Torres~Martins, Dipanjan Das, Noah~A. Smith, and Eric~P. Xing.
		\newblock 2008.
		\newblock Stacking dependency parsers.
		\newblock In {\em Proc. of the 2008 Conference on EMNLP}, pages 157--166,
		October.
		
		\bibitem[\protect\citename{Weiss \bgroup et al.\egroup
		}2015]{weiss-EtAl:2015:ACL-IJCNLP}
		David Weiss, Chris Alberti, Michael Collins, and Slav Petrov.
		\newblock 2015.
		\newblock Structured training for neural network transition-based parsing.
		\newblock In {\em Proc. of the 53rd ACL and the 7th IJCNLP (Volume 1: Long
			Papers)}, pages 323--333, July.
		
		\bibitem[\protect\citename{Zhang and Barzilay}2015]{zhang2015multi}
		Yuan Zhang and Regina Barzilay.
		\newblock 2015.
		\newblock Hierarchical low-rank tensors for multilingual transfer parsing.
		\newblock In {\em Proc. of the 2015 Conference on EMNLP}, pages 1857--1867,
		September.
		
		\bibitem[\protect\citename{Zhang and Clark}2008]{zhang-clark:2008:EMNLP}
		Yue Zhang and Stephen Clark.
		\newblock 2008.
		\newblock A tale of two parsers: {I}nvestigating and combining graph-based and
		transition-based dependency parsing.
		\newblock In {\em Proc. of the 2008 Conference on EMNLP}, pages 562--571,
		October.
		
		\bibitem[\protect\citename{Zhang \bgroup et al.\egroup
		}2014]{zhang-EtAl:2014:Coling2}
		Meishan Zhang, Wanxiang Che, Yanqiu Shao, and Ting Liu.
		\newblock 2014.
		\newblock Jointly or separately: Which is better for parsing heterogeneous
		dependencies?
		\newblock In {\em Proc. of COLING 2014}, pages 530--540, August.
		
		\bibitem[\protect\citename{Zhou \bgroup et al.\egroup
		}2015]{zhou-EtAl:2015:ACL-IJCNLP3}
		Hao Zhou, Yue Zhang, Shujian Huang, and Jiajun Chen.
		\newblock 2015.
		\newblock A neural probabilistic structured-prediction model for
		transition-based dependency parsing.
		\newblock In {\em Proc. of the 53rd ACL and the 7th IJCNLP (Volume 1: Long
			Papers)}, pages 1213--1222, July.
		
	\end{thebibliography}
	
\end{document}